\documentclass[10pt,twocolumn,letterpaper]{article}

\usepackage{iccv}
\usepackage{times}
\usepackage{epsfig}
\usepackage{graphicx}
\usepackage{amsmath}
\usepackage{amssymb}
\usepackage{subfigure}
\usepackage{diagbox}
\usepackage{overpic}

\newcommand{\G}{\mathcal{G}}
\newcommand{\V}{\mathcal{V}}
\newcommand{\E}{\mathcal{E}}

\newcommand{\R}{\mathbb{R}}

\usepackage{enumitem}
\newenvironment{tight_itemize}{
\begin{itemize}[leftmargin=20pt]
  \setlength{\topsep}{0pt}
  \setlength{\itemsep}{0pt}
  \setlength{\parskip}{0pt}
  \setlength{\parsep}{0pt}
}{\end{itemize}}

\newcommand*{\affaddr}[1]{#1} 
\newcommand*{\affmark}[1][*]{\textsuperscript{#1}}
\renewcommand\thefootnote{}

\usepackage[pagebackref=true,breaklinks=true,letterpaper=true,colorlinks,bookmarks=false]{hyperref}

\iccvfinalcopy 
\def\iccvPaperID{0000} 

\ificcvfinal\fi
\begin{document}

\title{RetinaFace: Single-stage Dense Face Localisation in the Wild}

\author{
Jiankang Deng \textsuperscript{*} \affmark[1,2,4] \qquad Jia Guo \textsuperscript{*} \affmark[2] \qquad Yuxiang Zhou \affmark[1] \qquad \\
Jinke Yu \affmark[2] \qquad Irene Kotsia \affmark[3] \qquad Stefanos Zafeiriou\affmark[1,4]\\
\affaddr{\affmark[1]Imperial College London} \qquad
\affaddr{\affmark[2]InsightFace} \qquad
\affaddr{\affmark[3]Middlesex University London} \qquad
\affaddr{\affmark[4]FaceSoft}\\
}

\maketitle

\begin{abstract}
Though tremendous strides have been made in uncontrolled face detection, accurate and efficient face localisation in the wild remains an open challenge. This paper presents a robust single-stage face detector, named RetinaFace, which performs pixel-wise face localisation on various scales of faces by taking advantages of joint extra-supervised and self-supervised multi-task learning. Specifically, We make contributions in the following five aspects: 
(1) We manually annotate five facial landmarks on the WIDER FACE dataset and observe significant improvement in hard face detection with the assistance of this extra supervision signal.
(2) We further add a self-supervised mesh decoder branch for predicting a pixel-wise 3D shape face information in parallel with the existing supervised branches. 
(3) On the WIDER FACE hard test set, RetinaFace outperforms the state of the art average precision (AP) by $1.1\%$ (achieving AP equal to {\bf $91.4\%$}).
(4) On the IJB-C test set, RetinaFace enables state of the art methods (ArcFace) to improve their results in face verification (TAR=$89.59\%$ for FAR=1e-6).
(5) By employing light-weight backbone networks, RetinaFace can run real-time on a single CPU core for a VGA-resolution image.
Extra annotations and code have been made available at: \url{https://github.com/deepinsight/insightface/tree/master/RetinaFace}.

\footnote{\textsuperscript{*} Equal contributions. \\
Email: j.deng16@imperial.ac.uk; guojia@gmail.com \\
InsightFace is a nonprofit Github project for 2D and 3D face analysis.} 
\setcounter{footnote}{0}
\renewcommand\thefootnote{\arabic{footnote}}

\end{abstract}

\section{Introduction}

Automatic face localisation is the prerequisite step of facial image analysis for many applications such as facial attribute (\eg expression~\cite{zhang2018jointexpression} and age~\cite{pan2018mean}) and facial identity recognition~\cite{schroff2015facenet,liu2017sphereface,wang2018cosface,deng2018arcface}. A narrow definition of face localisation may refer to traditional face detection~\cite{viola2004robust,zafeiriou2015survey}, which aims at estimating the face bounding boxes without any scale and position prior. Nevertheless, in this paper we refer to a broader definition of face localisation which includes face detection~\cite{ramanan2012face}, face alignment~\cite{feng2018wing}, pixel-wise face parsing~\cite{smith2013exemplar} and 3D dense correspondence regression~\cite{alp2017densereg,feng2018joint}. That kind of dense face localisation provides accurate facial position information for all different scales. 

\begin{figure}[t]
\centering
\includegraphics[width=0.5\textwidth]{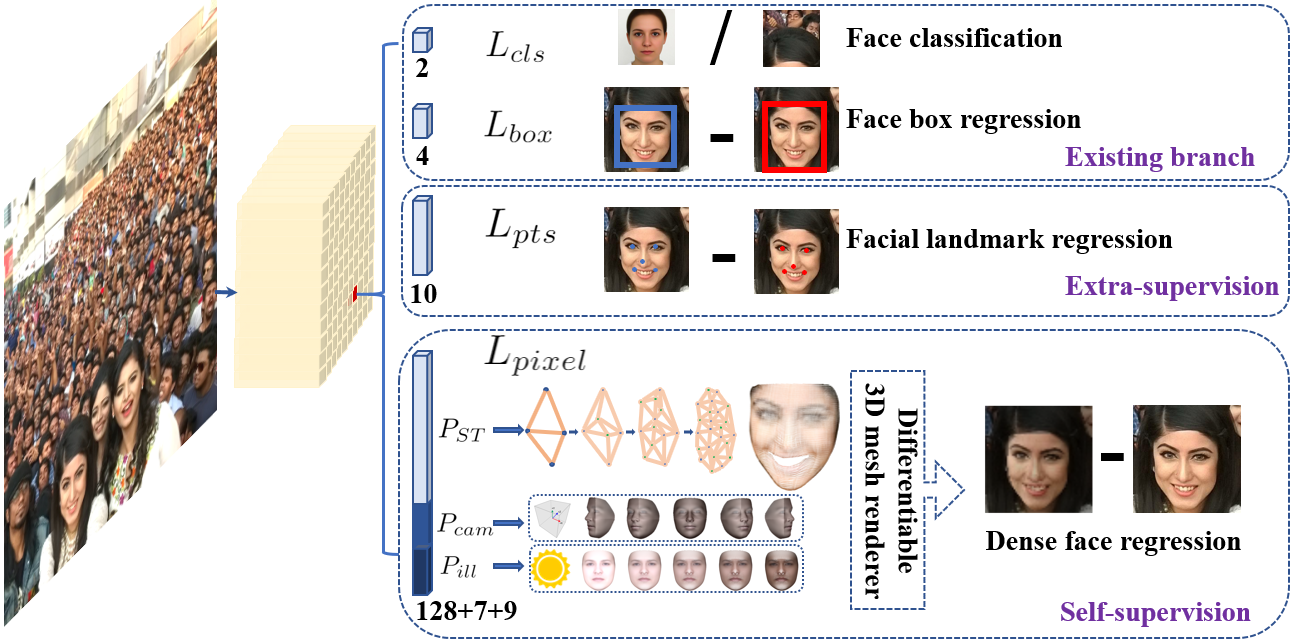}
\caption{The proposed single-stage pixel-wise face localisation method employs extra-supervised and self-supervised multi-task learning in parallel with the existing
box classification and regression branches. Each positive anchor outputs (1) a face score, (2) a face box, (3) five facial landmarks, and (4) dense 3D face vertices projected on the image plane.}
\label{fig:modelhead}
\vspace{-4mm}
\end{figure}

Inspired by generic object detection methods~\cite{girshick2015fast,ren2015faster,liu2016ssd,redmon2016you,redmon2017yolo9000,lin2017feature,lin2017focal}, which embraced all the recent advances in deep learning, face detection has recently achieved remarkable progress~\cite{hu2017finding,najibi2017ssh,zhang2017s3fd,chi2018selective,tang2018pyramidbox}. Different from generic object detection, face detection features smaller ratio variations (from 1:1 to 1:1.5) but much larger scale variations (from several pixels to thousand pixels). The most recent state-of-the-art methods~\cite{najibi2017ssh,zhang2017s3fd,tang2018pyramidbox} focus on single-stage~\cite{liu2016ssd,lin2017focal} design which densely samples face locations and scales on feature pyramids~\cite{lin2017feature},
demonstrating promising performance and yielding faster speed compared to two-stage methods~\cite{ren2015faster,zhangchangzheng2018face,chi2018selective}. Following  this route, we improve the single-stage face detection framework and propose a state-of-the-art dense face localisation method
by exploiting multi-task losses coming from strongly supervised and self-supervised signals. Our idea is examplified in Fig. \ref{fig:modelhead}.

Typically, face detection training process contains both classification and box regression losses ~\cite{girshick2015fast}.
Chen \etal~\cite{chen2014joint} proposed to combine face detection and alignment in a joint cascade framework based on the observation that aligned face shapes provide better features for face classification. Inspired by \cite{chen2014joint}, MTCNN~\cite{zhang2016joint} and STN~\cite{chen2016supervised}  simultaneously detected faces and five facial landmarks. 
Due to training data limitation, JDA~\cite{chen2014joint}, MTCNN~\cite{zhang2016joint} and STN~\cite{chen2016supervised} have not verified whether tiny face detection can benefit from the extra supervision of five facial landmarks. One of the questions we aim at answering in this paper is whether we can push forward the current best performance ($90.3\%$~\cite{zhang2019improved}) on the WIDER FACE hard test set~\cite{yang2016wider} by using extra supervision signal built of five facial landmarks. 

In Mask R-CNN~\cite{he2017mask}, the detection performance is significantly improved by adding a branch for predicting an object mask in parallel with the existing branch for bounding box recognition and regression. That confirms that dense pixel-wise annotations are also beneficial to improve detection. Unfortunately, for the challenging faces of WIDER FACE it is not possible to conduct dense face annotation (either in the form of more landmarks or semantic segments). Since supervised signals cannot be easily obtained, the question is whether we can apply unsupervised methods to further improve face detection. 

In FAN~\cite{wang2017faceattention}, an anchor-level attention map is proposed to improve the occluded face detection. Nevertheless, the proposed attention map is quite coarse and does not contain semantic information. Recently, self-supervised 3D morphable models~\cite{Genova2018CVPR,tran2018nonlinear,tran2018learning,zhou2019CVPR2500FPS} have achieved promising 3D face modelling in-the-wild. Especially, Mesh Decoder~\cite{zhou2019CVPR2500FPS} achieves over real-time speed by exploiting graph convolutions~\cite{defferrard2016convolutional,ranjan2018generating} on joint shape and texture. However, the main challenges of applying mesh decoder~\cite{zhou2019CVPR2500FPS} into the single-stage detector are: (1) camera parameters are hard to estimate accurately, and (2) the joint latent shape and texture representation is predicted from a single feature vector ($1\times1$ Conv on feature pyramid) instead of the RoI pooled feature, which indicates the risk of feature shift. In this paper, we employ a mesh decoder~\cite{zhou2019CVPR2500FPS} branch through self-supervision learning for predicting a pixel-wise 3D face shape in parallel with the existing supervised branches. 

To summarise, our key contributions are:
\vspace{-0.2cm}
\begin{tight_itemize}
\item Based on a single-stage design, we propose a novel pixel-wise face localisation method named RetinaFace, which employs a multi-task learning strategy to simultaneously predict face score, face box, five facial landmarks, and 3D position and correspondence of each facial pixel. 
\item On the WIDER FACE hard subset, RetinaFace outperforms the AP of the state of the art two-stage method (ISRN~\cite{zhang2019improved}) by $1.1\%$ (AP equal to $91.4\%$). 
\item On the IJB-C dataset, RetinaFace helps to improve ArcFace's~\cite{deng2018arcface} verification accuracy (with TAR equal to  $89.59\%$ when FAR=1e-6). This indicates that better face localisation can significantly improve face recognition.
\item By employing light-weight backbone networks, RetinaFace can run real-time on a single CPU core for a VGA-resolution image.
\item Extra annotations and code have been released to facilitate future research.
\end{tight_itemize}

\section{Related Work}

\noindent{\bf Image pyramid v.s. feature pyramid:} The sliding-window paradigm, in which a classifier is applied on a dense image grid, can be traced back to past decades. The milestone work of Viola-Jones~\cite{viola2004robust} explored cascade chain to reject false face regions from an image pyramid with real-time efficiency, leading to the widespread adoption of such scale-invariant face detection framework~\cite{zhang2016joint,chen2016supervised}. Even though the sliding-window on image pyramid was the leading detection paradigm~\cite{Hao2017CVPR,liu2017recurrent}, with the emergence of feature pyramid~\cite{lin2017feature}, sliding-anchor~\cite{ren2015faster} on multi-scale feature maps~\cite{zhang2017s3fd,tang2018pyramidbox}, quickly dominated face detection.

\noindent{\bf Two-stage v.s. single-stage:} Current face detection methods have inherited some achievements from generic object detection approaches and can be divided into two categories: two-stage methods (\eg Faster R-CNN~\cite{ren2015faster,zhangchangzheng2018face,zhu2017cms}) and single-stage methods (\eg SSD~\cite{liu2016ssd,zhang2017s3fd} and RetinaNet~\cite{lin2017focal,tang2018pyramidbox}). Two-stage methods employed a ``proposal and refinement'' mechanism featuring high localisation accuracy. By contrast, single-stage methods densely sampled face locations and scales, which resulted in extremely unbalanced positive and negative samples during training. To handle this imbalance, sampling~\cite{shrivastava2016training} and re-weighting~\cite{lin2017focal} methods were widely adopted. Compared to two-stage methods, single-stage methods are more efficient and have higher recall rate but at the risk of achieving a higher false positive rate and compromising the localisation accuracy. 

\noindent{\bf Context Modelling:} To enhance the model's contextual reasoning power for capturing tiny faces~\cite{hu2017finding}, 
SSH~\cite{najibi2017ssh} and PyramidBox~\cite{tang2018pyramidbox} applied context modules on feature pyramids to enlarge the receptive field from Euclidean grids. To enhance the non-rigid transformation modelling capacity of CNNs, deformable convolution network (DCN)~\cite{dai2017deformable,zhu2018deformable}
employed a novel deformable layer to model geometric transformations. The champion solution of the WIDER Face Challenge 2018~\cite{loy2019wider} indicates that rigid (expansion) and non-rigid (deformation) context modelling are complementary and orthogonal to improve the performance of face detection.

\noindent{\bf Multi-task Learning:} Joint face detection and alignment is widely used~\cite{chen2014joint,zhang2016joint,chen2016supervised} as aligned face shapes provide better features for face classification. In Mask R-CNN~\cite{he2017mask}, the detection performance was significantly improved by adding a branch for predicting an object mask in parallel with the existing branches. Densepose~\cite{alp2018densepose} adopted the architecture of Mask-RCNN to obtain dense part labels and coordinates within each of the selected regions. Nevertheless, the dense regression branch in~\cite{he2017mask,alp2018densepose} was trained by supervised learning. In addition, the dense branch was a small FCN applied to each RoI to predict a pixel-to-pixel dense mapping. 

\section{RetinaFace}

\begin{figure*}[t]
\centering
\includegraphics[width=0.95\textwidth]{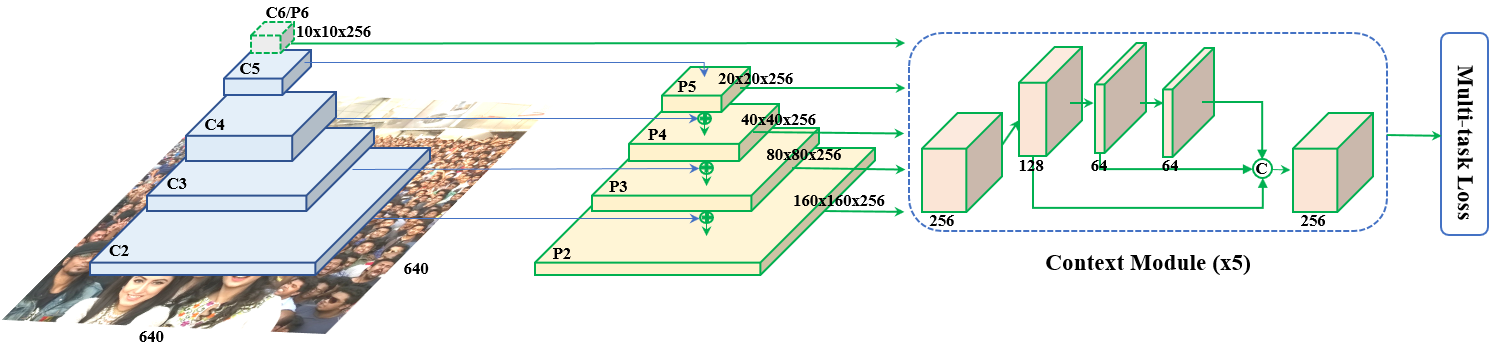}
\caption{An overview of the proposed single-stage dense face localisation approach. RetinaFace is designed based on the feature pyramids with independent context modules. Following the context modules, we calculate a multi-task loss for each anchor.}
\label{fig:framework}
\vspace{-4mm}
\end{figure*}

\subsection{Multi-task Loss}

For any training anchor $i$, we minimise the following multi-task loss:
\begin{equation}
\begin{split}
L  & =  L_{cls}(p_i, p^{*}_i) + \lambda_1 p^{*}_i L_{box}(t_i, t^{*}_i) \\
    & + \lambda_2 p^{*}_i L_{pts} (l_i, l^{*}_i) + \lambda_3 p^{*}_i L_{pixel}.\\
\end{split}
\label{eq:loss}
\end{equation}
(1) Face classification loss $L_{cls}(p_i, p^{*}_i)$, where $p_i$ is the predicted probability of anchor $i$ being a face and $p^{*}_i$ is 1 for the positive anchor and 0 for the negative anchor. The classification loss $L_{cls}$ is the softmax loss for binary classes (face/not face). 
(2) Face box regression loss $L_{box}(t_i, t^{*}_i)$, where $t_i=\{t_x, t_y, t_w, t_h\}_i$ and $t^{*}_i=\{t^{*}_x, t^{*}_y, t^{*}_w, t^{*}_h\}_i$ represent the coordinates of the predicted box and ground-truth box associated with the positive anchor. 
We follow~\cite{girshick2015fast} to normalise the box regression targets (\ie centre location, width and height) and use $L_{box}(t_i, t^{*}_i)=R(t_i - t^{*}_i)$, where $R$ is the robust loss function (smooth-L$_1$) defined in~\cite{girshick2015fast}. 
(3) Facial landmark regression loss $L_{pts} (l_i, l^{*}_i)$, where $l_i=\{l_{x_1}, l_{y_1}, \dots , l_{x_5}, l_{y_5}\}_i$ and $l^{*}_i=\{l^{*}_{x_1}, l^{*}_{y_1}, \dots , l^{*}_{x_5}, l^{*}_{y_5}\}_i$ represent the predicted five facial landmarks and ground-truth associated with the positive anchor. 
Similar to the box centre regression, the five facial landmark regression also employs the target normalisation based on the anchor centre.
(4) Dense regression loss $L_{pixel}$ (refer to Eq.~\ref{fig:selfsupervision}).  
The loss-balancing parameters $\lambda_1$-$\lambda_3$ are set to 0.25, 0.1 and 0.01, which means that we increase the significance of better box and landmark locations from supervision signals.

\begin{figure}[h]
\centering
\subfigure[2D Convolution]{
\label{fig:2Dconv}
\includegraphics[width=0.18\textwidth]{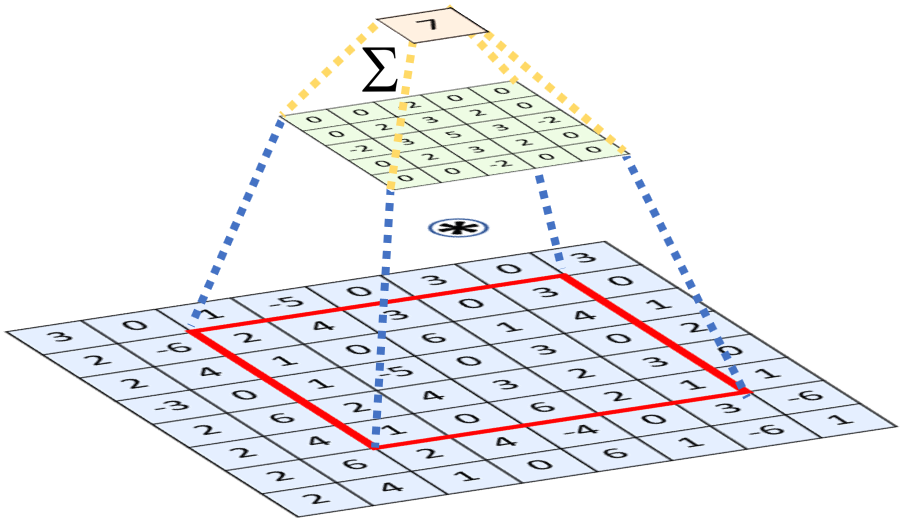}}
\subfigure[Graph Convolution]{
\label{fig:graphcon}
\includegraphics[width=0.25\textwidth]{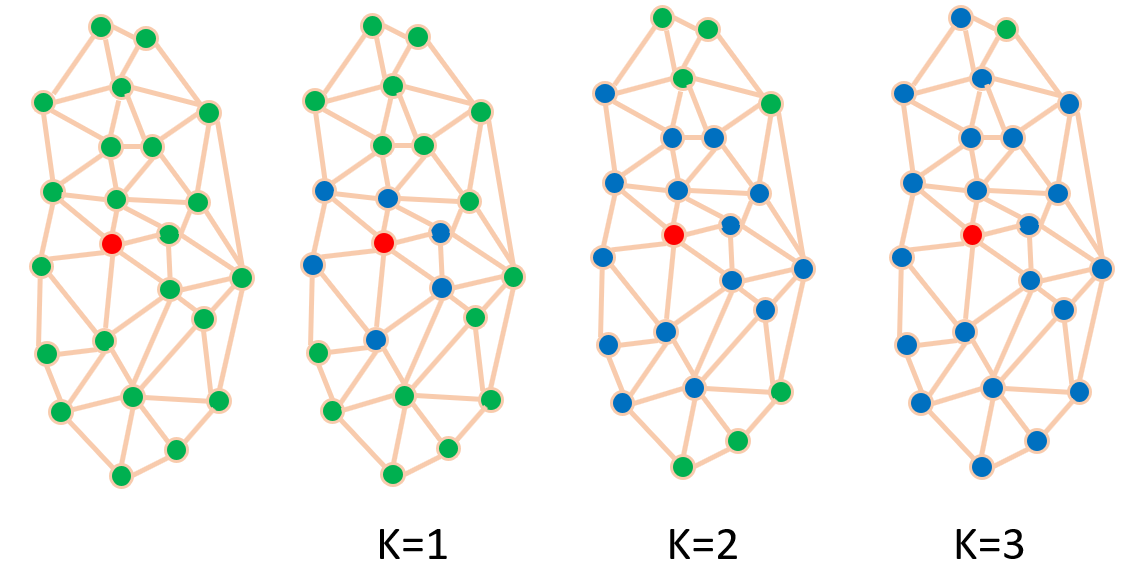}}
\caption{(a) 2D Convolution is kernel-weighted neighbour sum within the Euclidean grid receptive field. Each convolutional layer has $Kernel_H \times Kernel_W \times Channel_{in} \times Channel_{out}$ parameters. (b) Graph convolution is also in the form of kernel-weighted neighbour sum, but the neighbour distance is calculated on the graph by counting the minimum number of edges connecting two vertices. Each convolutional layer has $K \times Channel_{in} \times Channel_{out}$ parameters and the Chebyshev coefficients $\theta_{i,j} \in \mathbb{R}^K$ are truncated at order $K$.}
\label{fig:convcompare}
\vspace{-4mm}
\end{figure}

\subsection{Dense Regression Branch}

\noindent{\bf Mesh Decoder.} We directly employ the mesh decoder (mesh convolution and mesh up-sampling) from~\cite{zhou2019CVPR2500FPS,ranjan2018generating}, which is a graph convolution method based on fast localised spectral filtering~\cite{defferrard2016convolutional}. In order to achieve further acceleration, we also use a joint shape and texture decoder similarly to the method in \cite{zhou2019CVPR2500FPS}, contrary to \cite{ranjan2018generating} which only decoded shape. 

Below we will briefly explain the concept of graph convolutions and outline why they can be used for fast decoding. As illustrated in Fig.~\ref{fig:2Dconv}, a 2D convolutional operation is a ``kernel-weighted neighbour sum'' within the Euclidean grid receptive field. Similarly, graph convolution also employs the same concept as shown in Fig.~\ref{fig:graphcon}. 
However, the neighbour distance is calculated on the graph by counting the minimum number of edges connecting two vertices. We follow~\cite{zhou2019CVPR2500FPS} to define a coloured face mesh $\G=(\V,\E)$, where $\V \in \mathbb{R}^{n \times 6}$ is a set of face vertices containing the joint shape and texture information, and $\E \in \{0,1\}^{n \times n}$ is a {\bf sparse} adjacency matrix encoding the connection status between vertices. The graph Laplacian is defined as $L = D -\E \in\R^{n \times n}$ where $D \in \R^{n \times n}$ is a diagonal matrix with $D_{ii} = \sum_j \E_{ij}$. 

Following~\cite{defferrard2016convolutional,ranjan2018generating,zhou2019CVPR2500FPS}, the graph convolution with kernel $g_\theta$ can be formulated as a recursive Chebyshev polynomial truncated at order $K$,
\begin{equation}
y = g_\theta(L) x= \sum_{k=0}^{K-1} \theta_k T_k(\tilde{L}) x,
\label{chebyshevpolynomial}
\end{equation}
where $\theta \in \R^K$ is a vector of Chebyshev coefficients and $T_k(\tilde{L}) \in \R^{n\times n}$ is the Chebyshev polynomial of order $k$
evaluated at the scaled Laplacian $\tilde{L}$.
% = 2 L / \lambda_{max} - I_n$ ($I_n$ is a identity matrix).
Denoting $\bar{x}_k = T_k(\tilde{L})x \in \R^n$, we can recurrently compute $\bar{x}_k = 2\tilde{L} \bar{x}_{k-1} - \bar{x}_{k-2}$ with $\bar{x}_0 = x$ and $\bar{x}_1 = \tilde{L}x$. The whole filtering operation is extremely efficient including $K$ {\bf sparse} matrix-vector multiplications and one dense matrix-vector multiplication $y = g_\theta(L) x = [\bar{x}_0, \ldots, \bar{x}_{K-1}] \theta$. 

\noindent{\bf Differentiable Renderer.} After we predict the shape and texture parameters $P_{ST}\in \mathbb{R}^{128}$, we employ an efficient differentiable 3D mesh renderer~\cite{Genova2018CVPR} to project a coloured-mesh $\mathcal{D}_{P_{ST}}$ onto a 2D image plane with camera parameters $P_{cam}=[x_c,y_c,z_c,x_c',y_c',z_c',f_c]$ (\ie camera location, camera pose and focal length) and illumination parameters $P_{ill} = [x_l,y_l,z_l,r_l,g_l,b_l,r_a,g_a,b_a]$ (\ie location of point light source, colour values and colour of ambient lighting).

\noindent{\bf Dense Regression Loss.}
Once we get the rendered 2D face $\mathcal{R}(\mathcal{D}_{P_{ST}},P_{cam},P_{ill})$, we compare the pixel-wise difference of the rendered and the original 2D face using the following function: 
\begin{equation}
L_{pixel}  = \frac{1}{W*H} \sum_{i}^{W} \sum_{j}^{H}  \|     \mathcal{R}(\mathcal{D}_{P_{ST}},P_{cam},P_{ill})_{i,j} - I_{i,j}^*  \|_1,\\
\label{fig:selfsupervision}
\end{equation}
where $W$ and $H$ are the width and height of the anchor crop $I_{i,j}^*$, respectively.

\section{Experiments}

\subsection{Dataset}

The WIDER FACE dataset~\cite{yang2016wider} consists of $32,203$ images and $393,703$ face bounding boxes with a high degree of variability in scale, pose, expression, occlusion and illumination. 
The WIDER FACE dataset is split into training ($40\%$), validation ($10\%$) and testing ($50\%$) subsets by randomly sampling from $61$ scene categories. 
Based on the detection rate of EdgeBox~\cite{zitnick2014edge}, three levels of difficulty (\ie Easy, Medium and Hard) are defined by incrementally incorporating hard samples.  

\begin{figure}[h]
\centering
\includegraphics[width=0.4\textwidth]{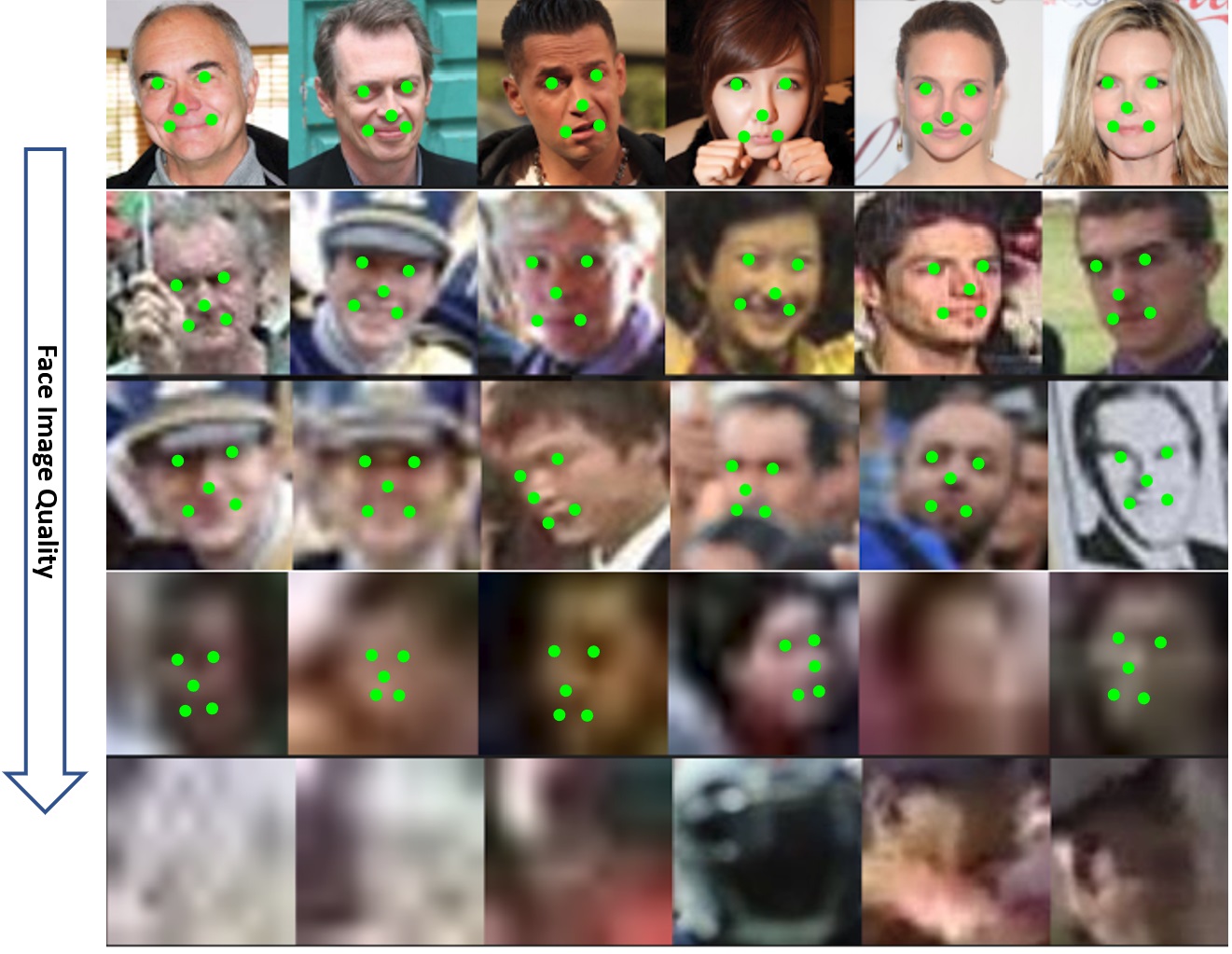}
\caption{We add extra annotations of five facial landmarks on faces that can be annotated (we call them ``annotatable'') from the WIDER FACE training and validation sets.}
\label{fig:extraanno}
\vspace{-4mm}
\end{figure}

\noindent{\bf Extra Annotations.} 
As illustrated in Fig.~\ref{fig:extraanno} and Tab.~\ref{tab:fivelevel}, we define five levels of face image quality (according to how difficult it is to annotate landmarks on the face) and annotate five facial landmarks (\ie eye centres, nose tip and mouth corners) on faces that can be annotated from the WIDER FACE training and validation subsets. In total, we have annotated $84.6k$ faces on the training set and $18.5k$ faces on the validation set.

\begin{table}[ht!]
\centering
\begin{tabular}{l|c|c}
\hline
Level            & Face Number &  Criterion\\
\hline
1 & 4,127  & indisputable 68 landmarks~\cite{sagonas2013300} \\
2 & 12,636 & annotatable 68 landmarks~\cite{sagonas2013300} \\
3 & 38,140 & indisputable 5 landmarks\\
4 & 50,024 & annotatable 5 landmarks\\
5 & 94,095 & distinguish by context  \\
\hline
\end{tabular}
\hspace{1in}
\caption{Five levels of face image quality. In the indisputable category a human can, without a lot of effort, locale the landmarks. In the annotatable category  finding an approximate location requires some effort.}
\label{tab:fivelevel}
\vspace{-4mm}
\end{table}

\subsection{Implementation details}

\noindent{\bf Feature Pyramid.} RetinaFace employs feature pyramid levels from $P_2$ to $P_6$, where $P_2$ to $P_5$ are computed from the output of the corresponding ResNet residual stage ($C_2$ through $C_5$) using top-down and lateral connections as in~\cite{lin2017feature,lin2017focal}. $P_6$ is calculated through a $3 \times 3$ convolution with stride=2 on $C_5$.
$C_1$ to $C_5$ is a pre-trained ResNet-152~\cite{he2016deep} classification network on the ImageNet-11k dataset while $P_6$ are randomly initialised with the ``Xavier" method~\cite{glorot2010understanding}. 

\noindent{\bf Context Module.} Inspired by SSH~\cite{najibi2017ssh} and PyramidBox~\cite{tang2018pyramidbox}, we also apply independent context modules on five feature pyramid levels to increase the receptive field and enhance the rigid context modelling power. Drawing lessons from the champion of the WIDER Face Challenge 2018~\cite{loy2019wider}, we also replace all $3\times3$ convolution layers within the lateral connections and context modules by the deformable convolution network (DCN)~\cite{dai2017deformable,zhu2018deformable}, which further strengthens the non-rigid context modelling capacity.

\noindent{\bf Loss Head.}
For negative anchors, only classification loss is applied. For positive anchors, the proposed multi-task loss is calculated. We employ a shared loss head ($1\times1$ conv) across different feature maps $H_n \times W_n \times 256, n\in \left \{ 2, \dots, 6 \right \}$. For the mesh decoder, we apply the pre-trained model~\cite{zhou2019CVPR2500FPS}, which is a small computational overhead that allows for efficient inference.

\noindent{\bf Anchor Settings.} As illustrated in Tab.~\ref{tab:anchors}, we employ scale-specific anchors on the feature pyramid levels from $P_2$ to $P_6$ like \cite{wang2017faceattention}. Here, $P_2$ is designed to capture tiny faces by tiling small anchors at the cost of more computational time and at the risk of more false positives. We set the scale step at $ 2 ^ {1/3}$ and the aspect ratio at 1:1. With the input image size at $640\times640$, the anchors can cover scales from $ 16\times16 $ to $ 406\times406 $ on the feature pyramid levels. In total, there are 102,300 anchors, and $75\%$ of these anchors are from $P_2$. 

\begin{table}[ht!]
\centering
\begin{tabular}{c|c|c}
\hline
Feature Pyramid & Stride & Anchor \\%& Number \\
\hline
$P_2$ ($160\times 160 \times 256$)& 4  & 16, 20.16, 25.40 \\%& 76800\\
$P_3$ ($80\times 80 \times 256$)  & 8  & 32, 40.32, 50.80 \\%& 19200\\
$P_4$ ($40\times 40 \times 256$)  & 16 & 64, 80.63, 101.59\\%& 4800\\
$P_5$ ($20\times 20 \times 256$)  & 32 & 128, 161.26, 203.19\\% & 1200\\
$P_6$ ($10\times 10 \times 256$)  & 64 & 256, 322.54, 406.37\\% & 300\\
\hline
\end{tabular}
\hspace{1in}
\caption{The details of feature pyramid, stride size, anchor in RetinaFace. For a $640\times640$ input image, there are 102,300 anchors in total, and $75\%$ of these anchors are tiled on $P_2$.}
\label{tab:anchors}
\vspace{-4mm}
\end{table}

During training, anchors are matched to a ground-truth box when IoU is larger than $0.5$, and to the background when IoU is less than $0.3$. Unmatched anchors are ignored during training. Since most of the anchors ($>99\%$) are negative after the matching step, we employ standard OHEM~\cite{shrivastava2016training,zhang2017s3fd} to alleviate significant imbalance between the positive and negative training examples. More specifically, we sort negative anchors by the loss values and select the top ones so that the ratio between the negative and positive samples is at least $3$:$1$. 

\noindent{\bf Data Augmentation.}
Since there are around $20\%$ tiny faces in the WIDER FACE training set, we follow~\cite{zhang2017s3fd,tang2018pyramidbox} and randomly crop square patches from the original images and resize these patches into $640\times640$ to generate larger training faces.
More specifically, square patches are cropped from the original image with a random size between [$0.3$, $1$] of the short edge of the original image. For the faces on the crop boundary, we keep the overlapped part of the face box if its centre is within the crop patch. Besides random crop, we also augment training data by random horizontal flip with the probability of $0.5$ and photo-metric colour distortion~\cite{zhang2017s3fd}.

\noindent{\bf Training Details.}
We train the RetinaFace using SGD optimiser (momentum at $0.9$, weight decay at $0.0005$, batch size of $8\times4$) on four NVIDIA Tesla P40 (24GB) GPUs. The learning rate starts from $10^{-3}$, rising to $10^{-2}$ after 5 epochs, then divided by $10$ at $55$ and $68$ epochs. The training process terminates at $80$ epochs. 

\noindent{\bf Testing Details.} For testing on WIDER FACE, we follow the standard practices of ~\cite{najibi2017ssh,zhang2017s3fd} and employ flip as well as multi-scale (the short edge of image at $[500, 800, 1100, 1400, 1700]$) strategies. Box voting~\cite{gidaris2015object} is applied on the union set of predicted face boxes using an IoU threshold at 0.4.

\subsection{Ablation Study}

To achieve a better understanding of the proposed RetinaFace, we conduct extensive ablation experiments to examine how the annotated five facial landmarks and the proposed dense regression branch quantitatively affect the performance of face detection. Besides the standard evaluation metric of average precision (AP) when IoU=0.5 on the Easy, Medium and Hard subsets, we also make use of the development server (Hard validation subset) of the WIDER Face Challenge 2018~\cite{loy2019wider}, which employs a more strict evaluation metric of mean AP (mAP) for IoU=0.5:0.05:0.95, rewarding more accurate face detectors.

As illustrated in Tab.~\ref{tab:ablationstudy}, we evaluate the performance of several different settings on the WIDER FACE validation set and focus on the observations of AP and mAP on the Hard subset. By applying the practices of state-of-the-art techniques (\ie FPN, context module, and deformable convolution), we set up a strong baseline ($91.286\%$), which is slightly better than ISRN~\cite{zhang2019improved} ($90.9\%$). Adding the branch of five facial landmark regression significantly improves the face box AP ($0.408\%$) and mAP ($0.775\%$) on the Hard subset, suggesting that landmark localisation is crucial for improving the accuracy of face detection. By contrast, adding the dense regression branch increases the face box AP on Easy and Medium subsets but slightly deteriorates the results on the Hard subset, indicating the difficulty of dense regression under challenging scenarios. 
Nevertheless, learning landmark and dense regression jointly enables a further improvement compared to adding landmark regression only. This demonstrates that landmark regression does help dense regression, which in turn boosts face detection performance even further.

\begin{table}[ht!]
\centering
\begin{tabular}{l|ccc|c}
\hline
Method  &Easy & Medium & Hard & mAP~\cite{loy2019wider}\\
\hline
FPN+Context & 95.532 & 95.134 & 90.714 & 50.842 \\
+DCN& 96.349 & 95.833 & 91.286 & 51.522\\ 
\hline
+$L_{pts}$ & 96.467 & 96.075 & 91.694 & 52.297\\ 
+$L_{pixel}$ & 96.413 & 95.864 & 91.276 & 51.492 \\ 
+$L_{pts}+L_{pixel}$ & {\bf 96.942} & {\bf 96.175} & {\bf 91.857} & {\bf 52.318}\\
\hline
\end{tabular}
\hspace{1in}
\caption{Ablation experiments of the proposed methods on the WIDER FACE validation subset.}
\label{tab:ablationstudy}
\vspace{-4mm}
\end{table}

\subsection{Face box Accuracy}

\begin{figure*}[t]
\centering
\subfigure[Val: Easy]{
\label{fig:ve}
\includegraphics[width=0.32\linewidth]{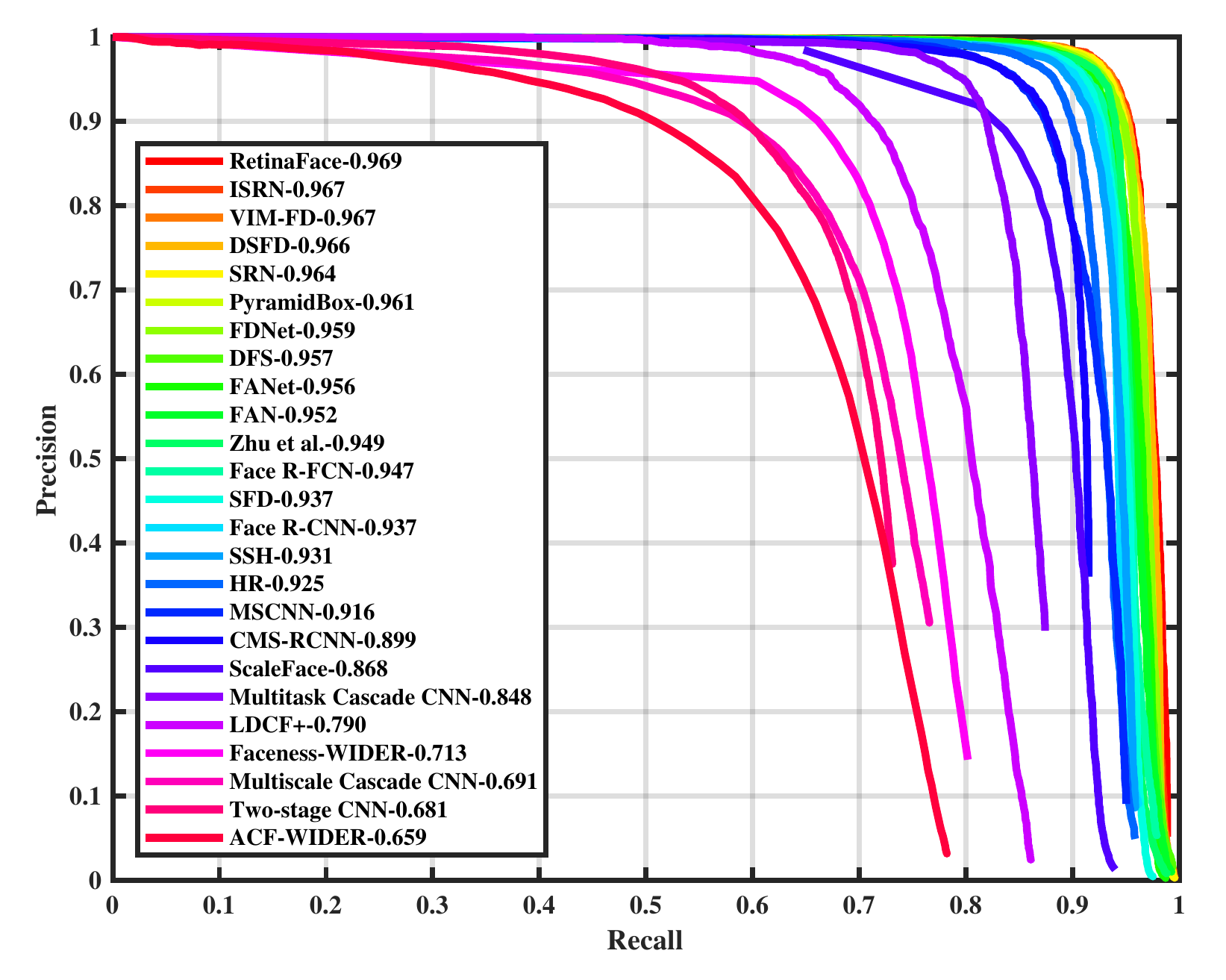}}
\subfigure[Val: Medium]{
\label{fig:vm}
\includegraphics[width=0.32\linewidth]{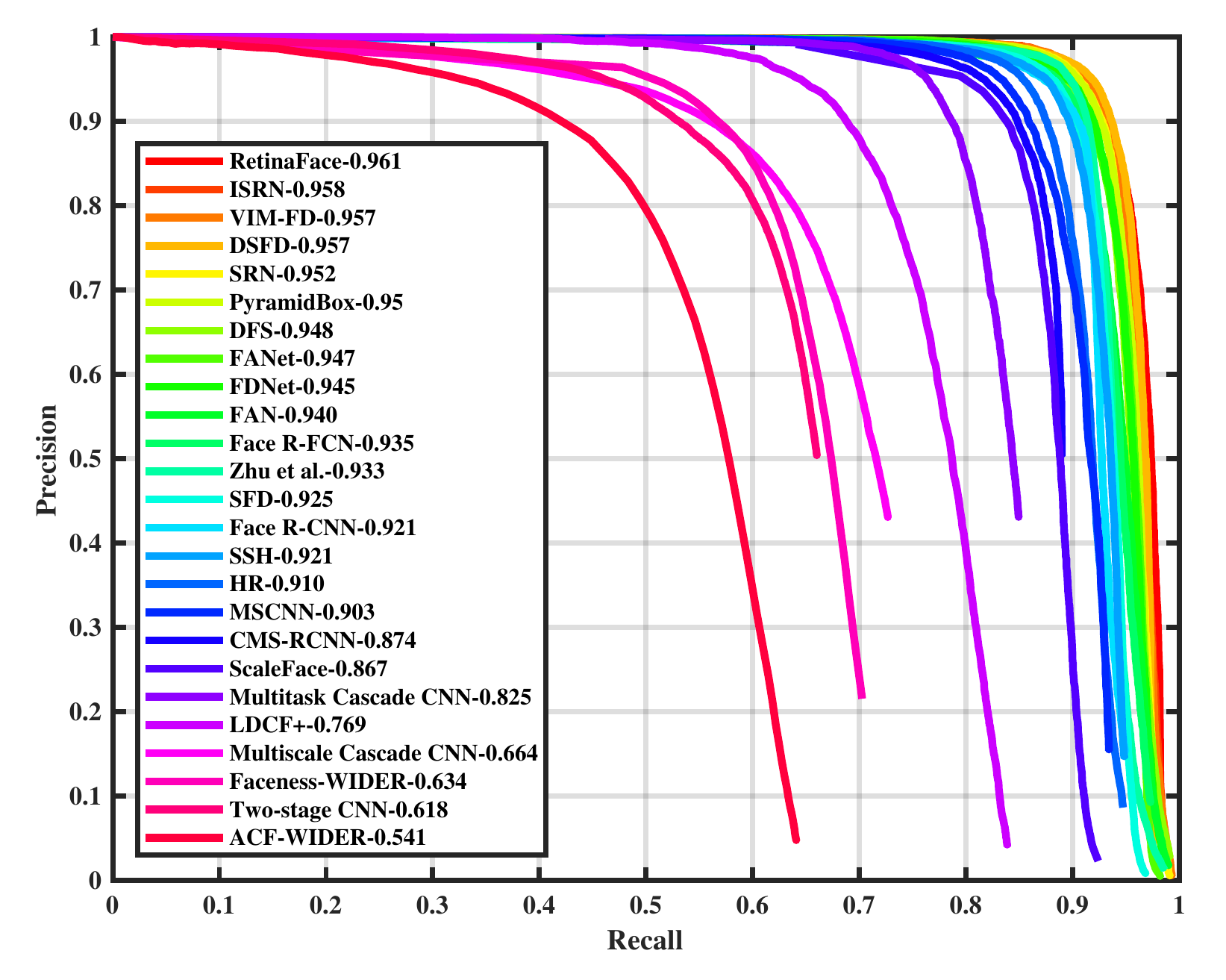}}
\subfigure[Val: Hard]{
\label{fig:vh}
\includegraphics[width=0.32\linewidth]{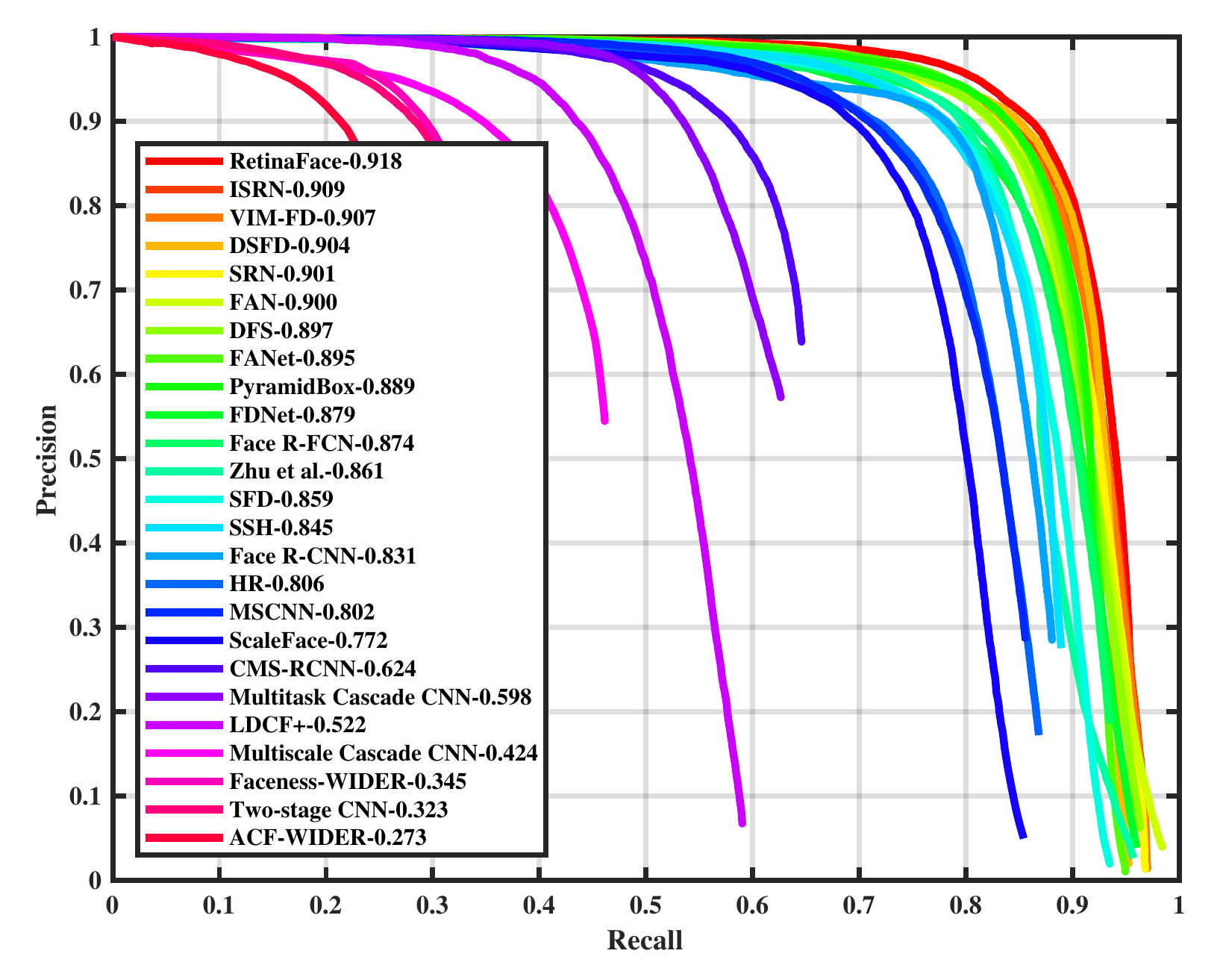}}
\subfigure[Test: Easy]{
\label{fig:te}
\includegraphics[width=0.32\linewidth]{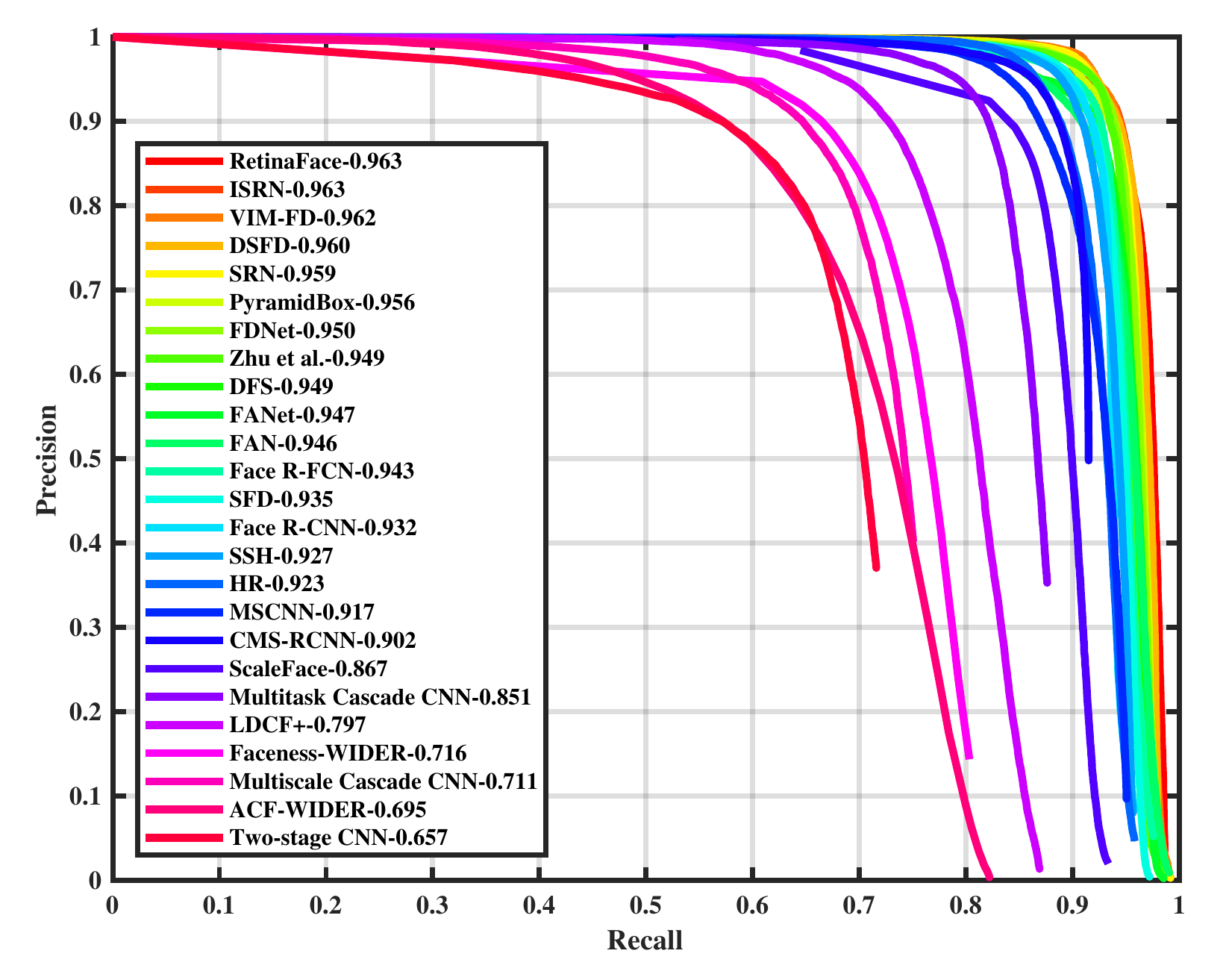}}
\subfigure[Test: Medium]{
\label{fig:tm}
\includegraphics[width=0.32\linewidth]{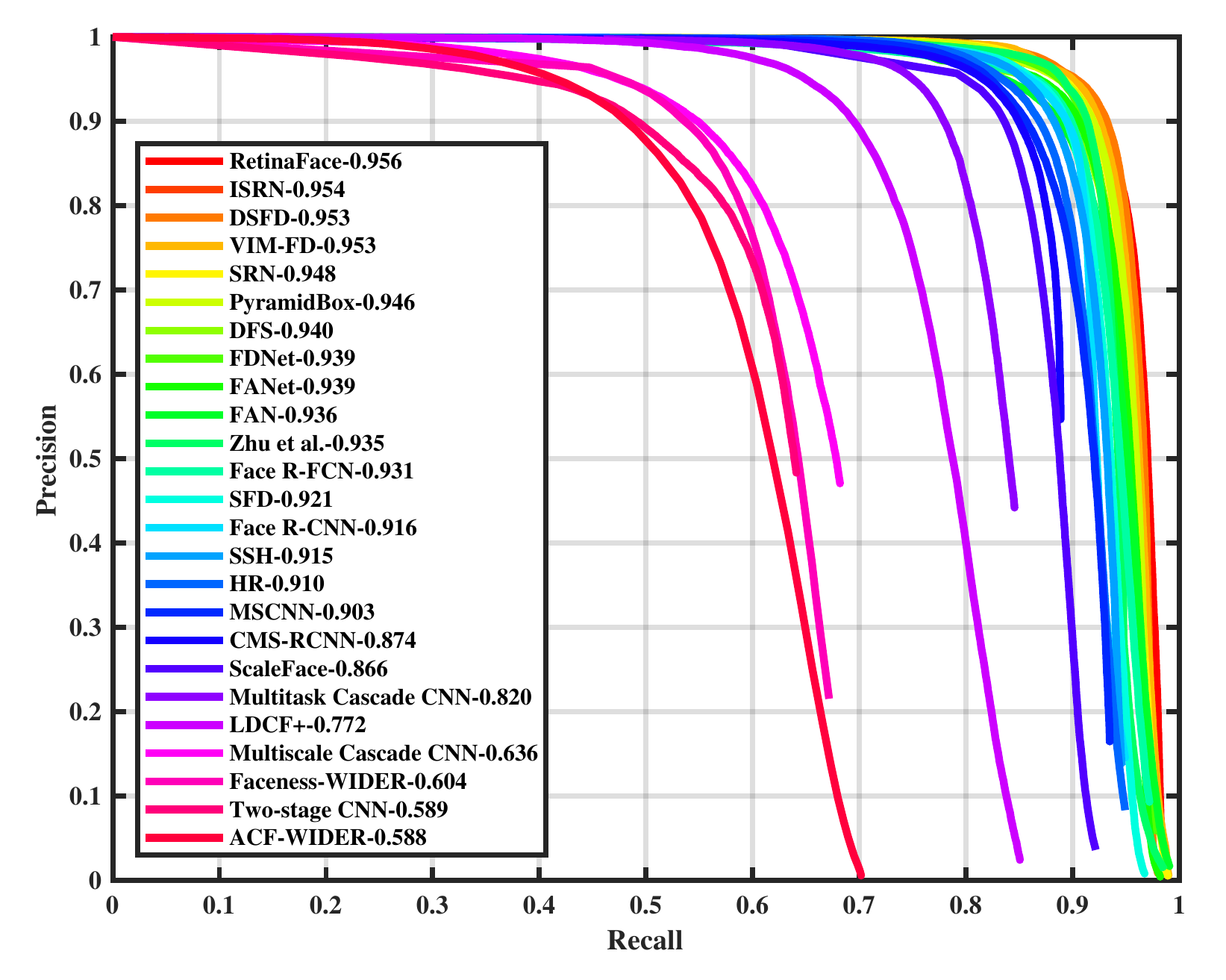}}
\subfigure[Test: Hard]{
\label{fig:th}
\includegraphics[width=0.32\linewidth]{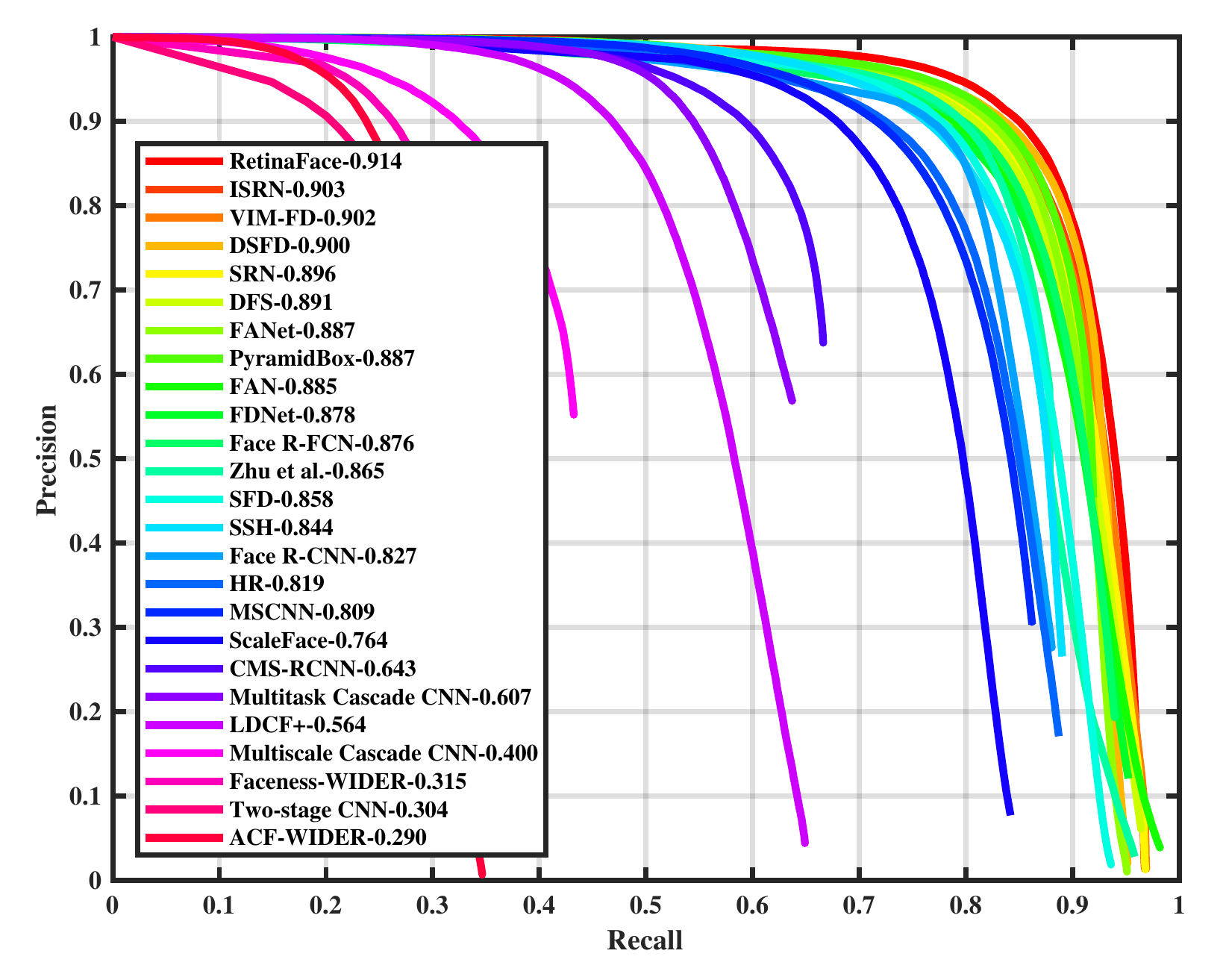}}
\caption{Precision-recall curves on the WIDER FACE validation and test subsets.}
\label{fig:wider-face}
\end{figure*}

Following the stander evaluation protocol of the WIDER FACE dataset, we only train the model on the training set and test on both the validation and test sets. To obtain the evaluation results on the test set, we submit the detection results to the organisers for evaluation. 
As shown in Fig.~\ref{fig:wider-face}, we compare the proposed RetinaFace with other $24$ state-of-the-art face detection algorithms 
(\ie Multiscale Cascade CNN~\cite{yang2016wider}, Two-stage CNN~\cite{yang2016wider}, ACF-WIDER~\cite{yang2014aggregate}, Faceness-WIDER~\cite{yang2015facial},
Multitask Cascade CNN~\cite{zhang2016joint}, CMS-RCNN~\cite{zhu2017cms}, LDCF+~\cite{ohn2016boost}, HR~\cite{hu2017finding}, Face R-CNN~\cite{wang2017face}, 
ScaleFace~\cite{yang2017face}, SSH~\cite{najibi2017ssh}, SFD~\cite{zhang2017s3fd}, Face R-FCN~\cite{wang2017detecting}, MSCNN~\cite{cai2016unified}, 
FAN~\cite{wang2017faceattention}, Zhu \etal~\cite{zhu2018seeing}, PyramidBox~\cite{tang2018pyramidbox}, FDNet~\cite{zhangchangzheng2018face}, SRN~\cite{chi2018selective},
FANet~\cite{zhang2017feature}, DSFD~\cite{li2018dsfd}, DFS~\cite{tian2018learning}, VIM-FD~\cite{zhang2019robust}, ISRN~\cite{zhang2019improved}). Our approach outperforms these state-of-the-art methods in terms of AP.
More specifically, RetinaFace produces the best AP in all subsets of both validation and test sets, \ie, $96.9\%$ (Easy), $96.1\%$ (Medium) and $91.8\%$ (Hard) for validation set, and $96.3\%$ (Easy), $95.6\%$ (Medium) and $91.4\%$ (Hard) for test set. Compared to the recent best performed method~\cite{zhang2019improved}, RetinaFace sets up a new impressive record ($91.4\%$ v.s. $90.3\%$) on the Hard subset which contains a large number of tiny faces.

\begin{figure*}[htp!]
\centering
\includegraphics[width=1.0\linewidth]{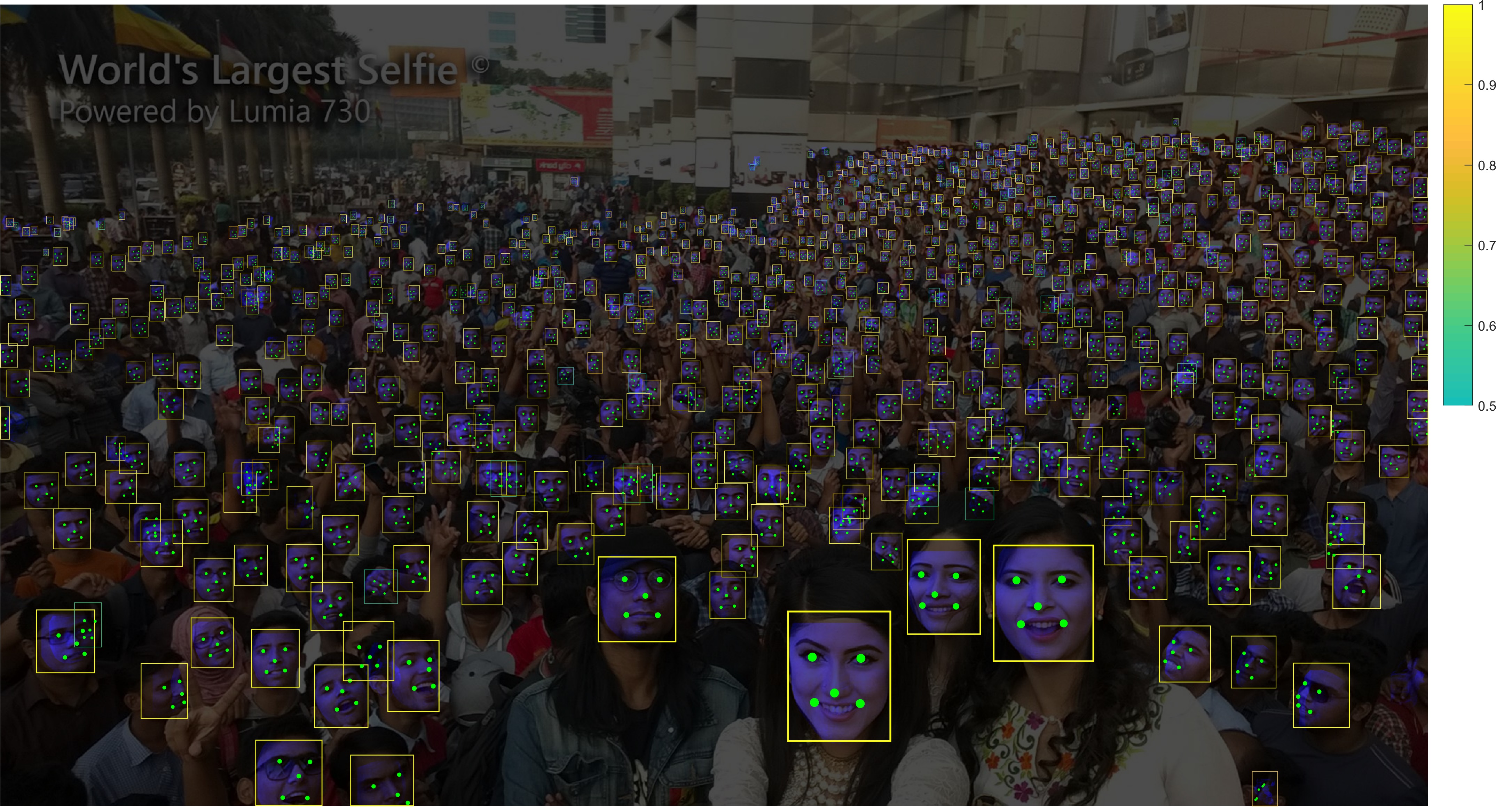}
\caption{RetinaFace can find around 900 faces (threshold at 0.5) out of the reported 1151 people, by taking advantages of the proposed joint extra-supervised and self-supervised multi-task learning. Detector confidence is given by the colour bar on the right. Dense localisation masks are drawn in blue. Please zoom in to check the detailed detection, alignment and dense regression results on tiny faces.}
\label{fig:impressivedemo}
\end{figure*}

% \noindent{\bf Qualitative Results.} 
In Fig.~\ref{fig:impressivedemo}, we illustrate qualitative results on a selfie with dense faces.
RetinaFace successfully finds about $900$ faces (threshold at 0.5) out of the reported $1,151$ faces.
Besides accurate bounding boxes, the five facial landmarks predicted by RetinaFace are also very robust under the variations of pose, occlusion and resolution. Even though there are some failure cases of dense face localisation under heavy occlusion, the dense regression results on some clear and large faces are good and even show expression variations. 
% \footnote{https://www.ndtv.com/offbeat/dont-miss-is-this-the-worlds-largest-selfie-703764}

\subsection{Five Facial Landmark Accuracy}

\begin{figure}[t]
\centering
\subfigure[NME on AFLW]{
\label{fig:5tpsaflw}
\includegraphics[height=0.16\textwidth]{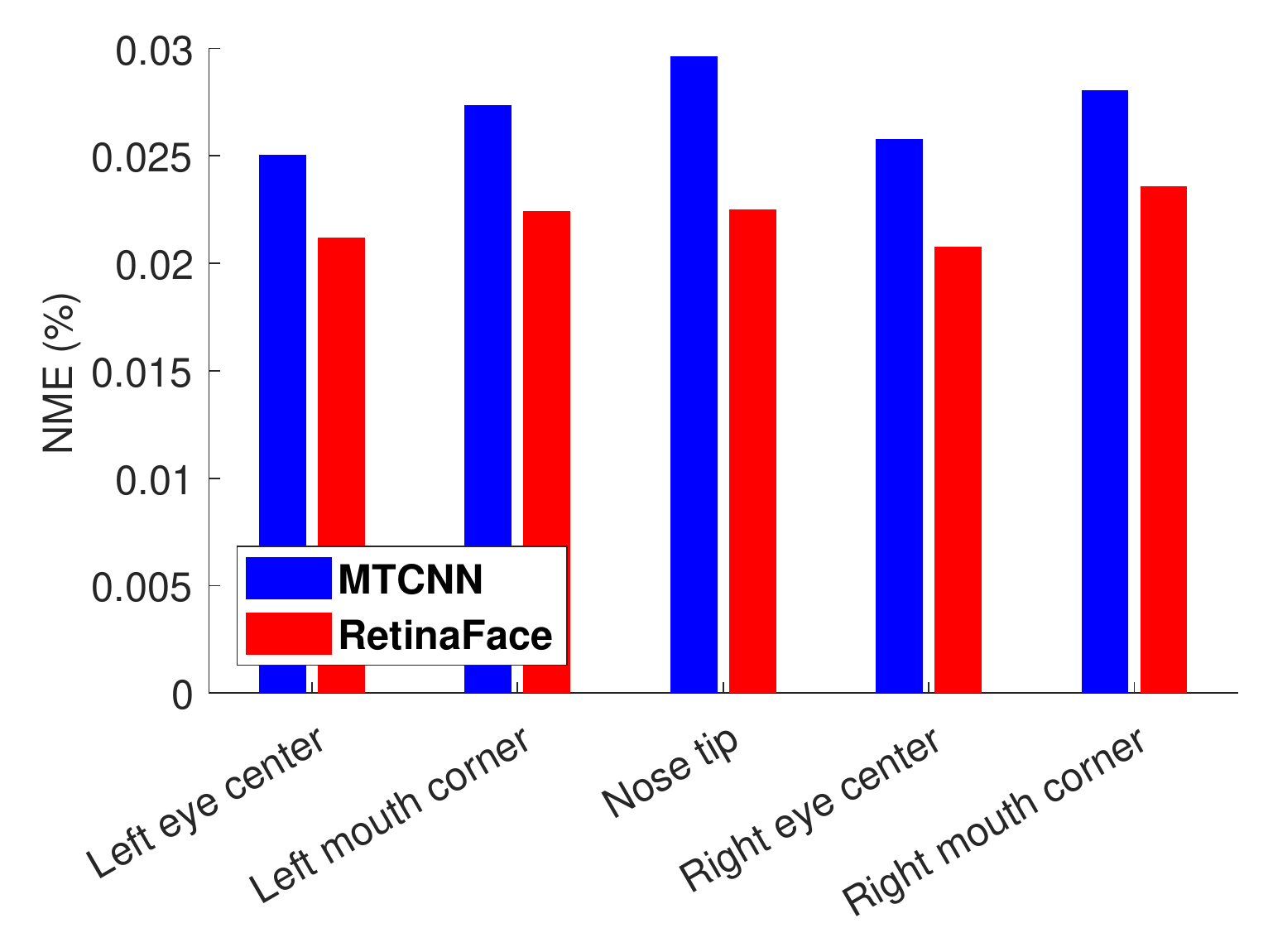}}
\subfigure[CED on WIDER FACE]{
\label{fig:5ptswider}
\includegraphics[height=0.16\textwidth]{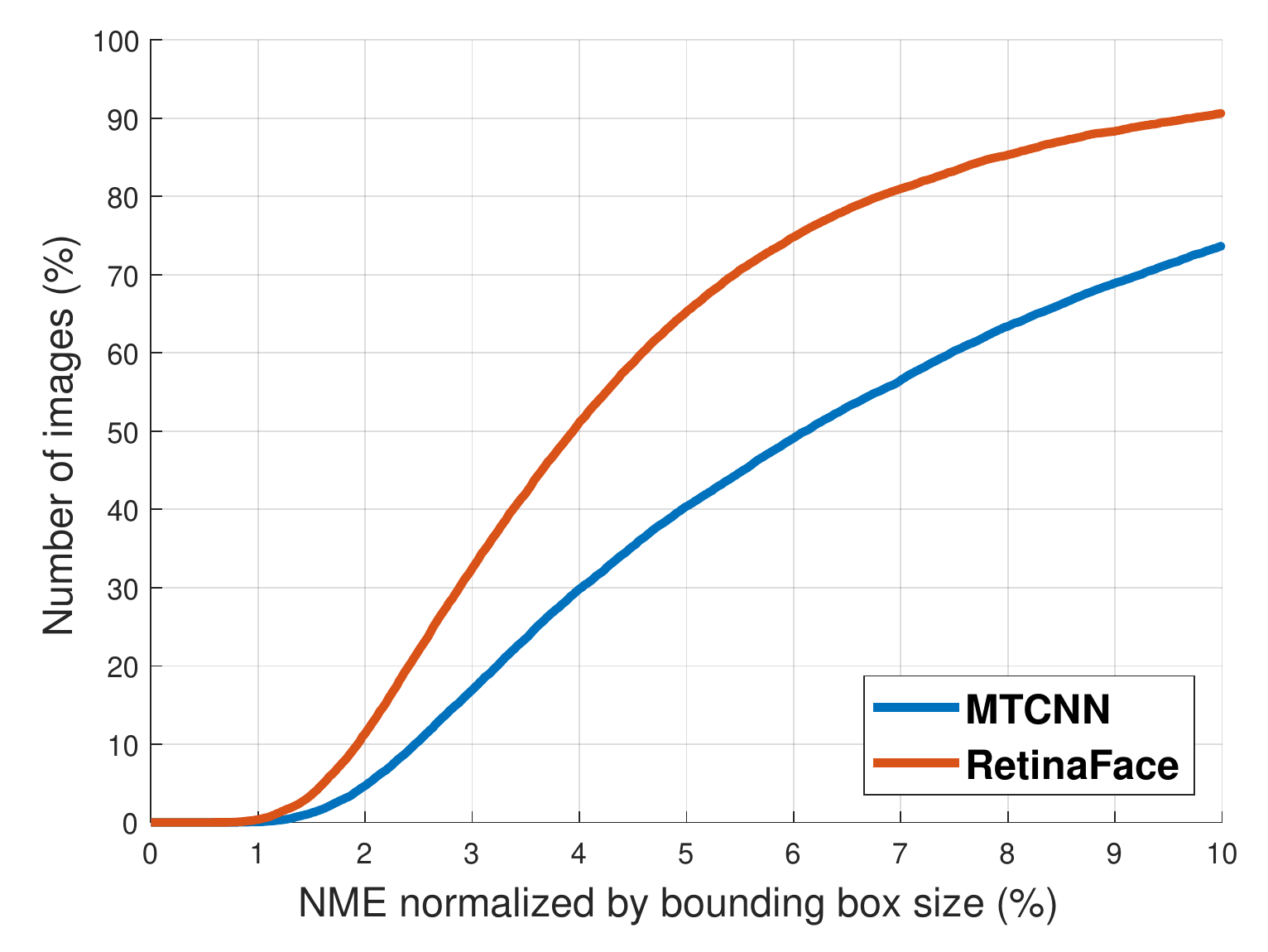}}
\caption{Qualitative comparison between MTCNN and RetinaFace on five facial landmark localisation. (a) AFLW (b) WIDER FACE validation set.
}
\label{fig:fiveptsresults}
\vspace{-4mm}
\end{figure}

To evaluate the accuracy of five facial landmark localisation, we compare RetinaFace with MTCNN~\cite{zhang2016joint} on the AFLW dataset~\cite{koestinger2011annotated} (24,386 faces) as well as the WIDER FACE validation set (18.5k faces). Here, we employ the face box size ($\sqrt{W\times H}$) as the normalisation distance. 
As shown in Fig.~\ref{fig:5tpsaflw}, we give the mean error of each facial landmark on the AFLW dataset~\cite{zhu2016unconstrained}. RetinaFace significantly decreases the normalised mean errors (NME) from $2.72\%$ to $2.21\%$ when compared to MTCNN. 
In Fig.~\ref{fig:5ptswider}, we show the cumulative error distribution (CED) curves on the WIDER FACE validation set. Compared to MTCNN, RetinaFace significantly decreases the failure rate from $26.31\%$ to $9.37\%$ (the NME threshold at $10\%$). 

\subsection{Dense Facial Landmark Accuracy}

\begin{figure}[ht!]
\centering
\subfigure[68 2D Landmarks]{
\label{fig:AFLW200068}
\includegraphics[width=0.23\textwidth]{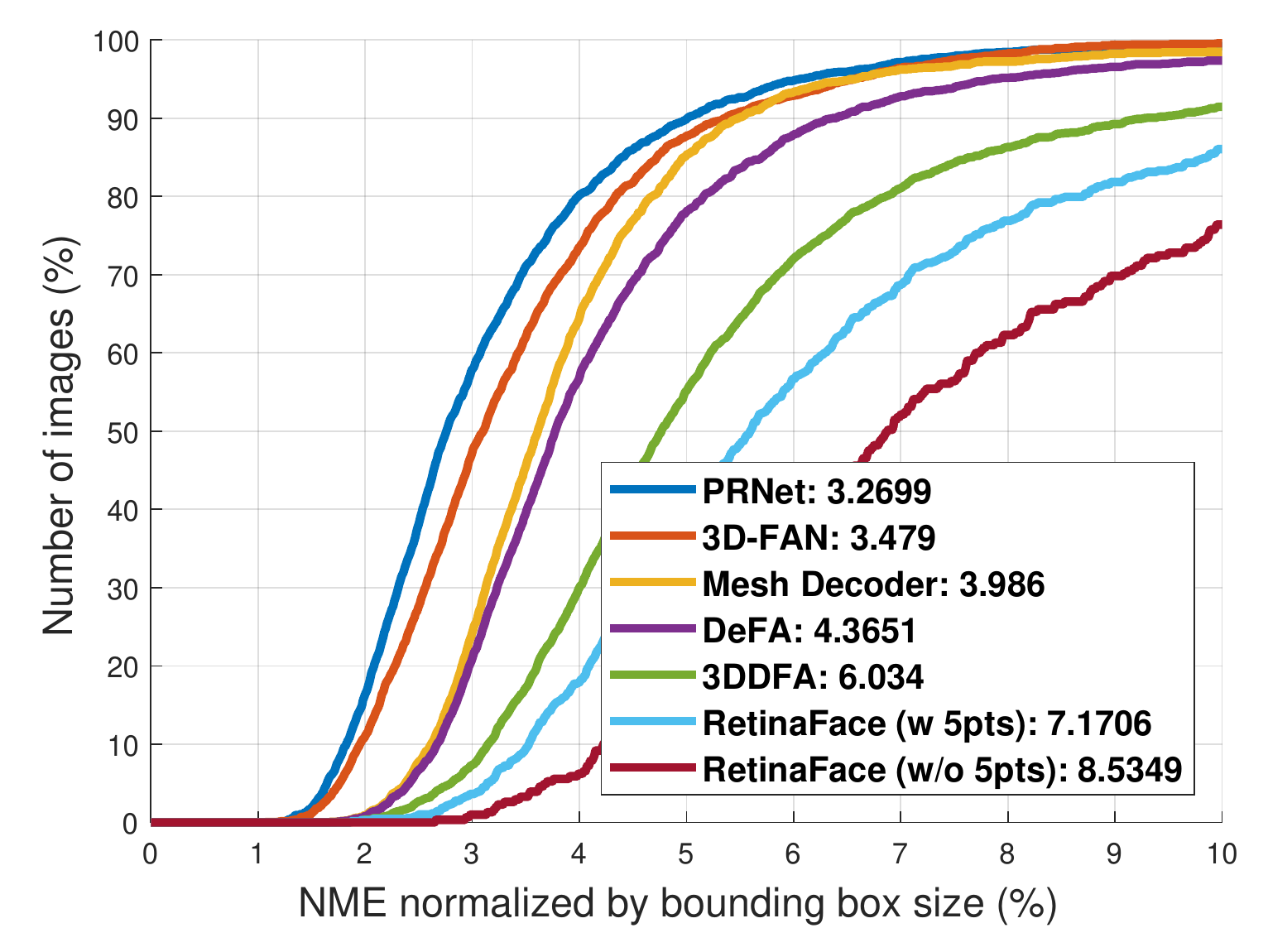}}
\subfigure[All 3D Landmarks]{
\label{fig:AFLW2000all}
\includegraphics[width=0.23\textwidth]{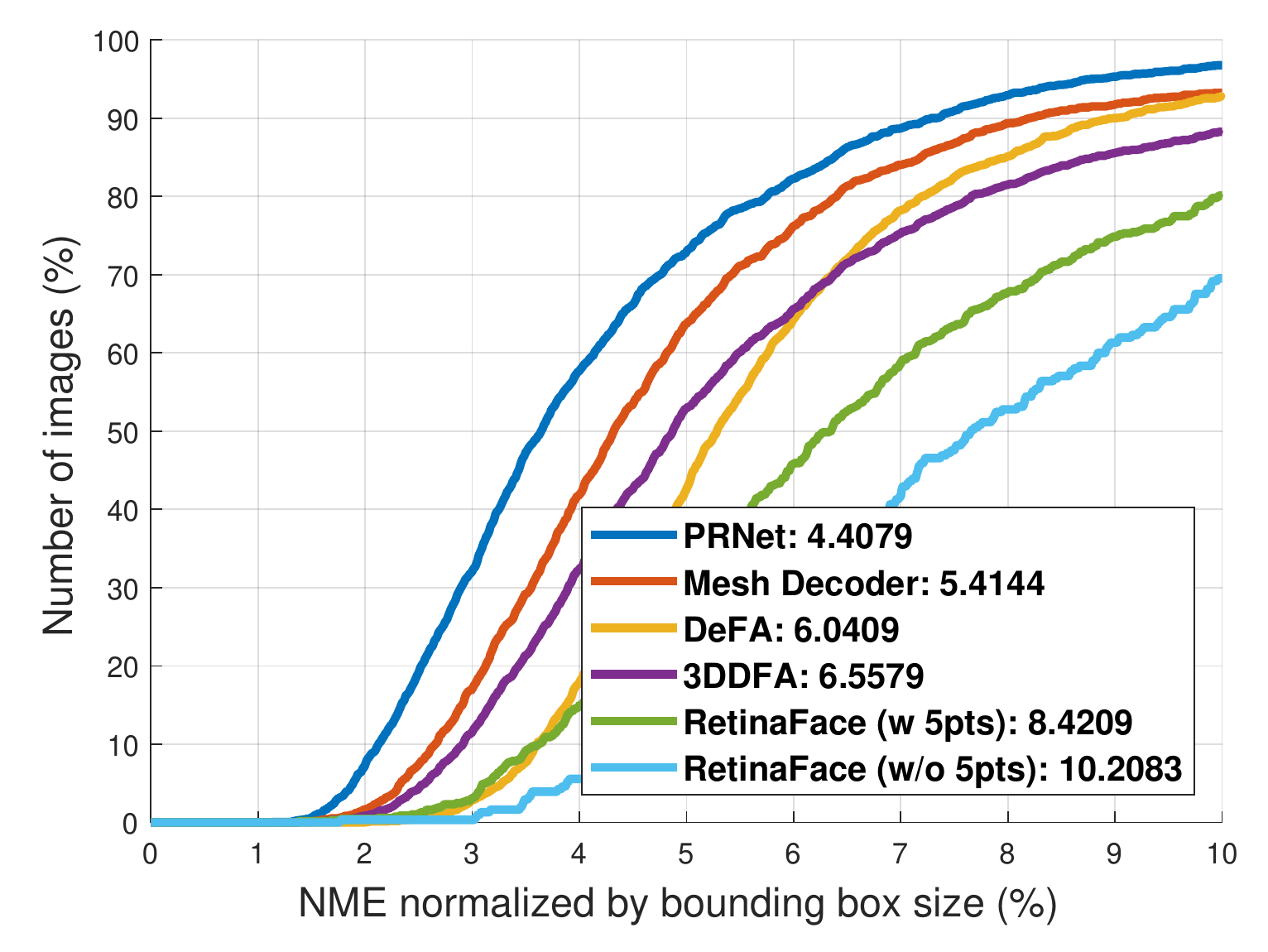}}
\subfigure[Result Analysis (Upper: Mesh Decoder;
Lower: RetinaFace)]{
\label{fig:denseanalysis}
\includegraphics[width=0.46\textwidth]{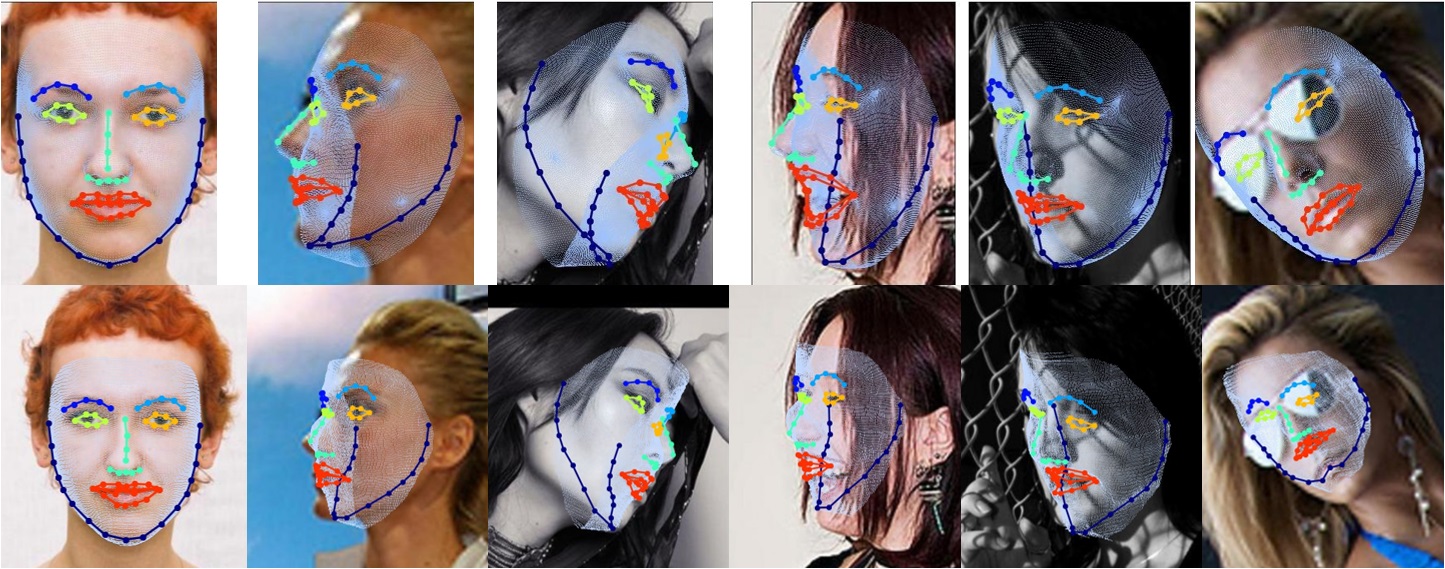}}
\caption{CED curves on AFLW2000-3D. Evaluation is performed on (a) 68 landmarks with the 2D coordinates and (b) all landmarks with 3D coordinates. In (c), we compare the dense regression results from RetinaFace and Mesh Decoder~\cite{zhou2019CVPR2500FPS}. RetinaFace can easily handle faces with pose variations but has difficulty to predict accurate dense correspondence under complex scenarios.}
\label{fig:AFLW2000-3D}
\vspace{-4mm}
\end{figure}

Besides box and five facial landmarks, RetinaFace also outputs dense face correspondence, but the dense regression branch is trained by self-supervised learning only.
Following~\cite{feng2018joint,zhou2019CVPR2500FPS}, we evaluate the accuracy of dense facial landmark localisation on the AFLW2000-3D dataset~\cite{zhu2016face} considering  (1) 68 landmarks with the 2D projection coordinates and (2) all landmarks with 3D coordinates. Here, the mean error is still normalised by the bounding box size~\cite{zhu2016face}. In Fig.~\ref{fig:AFLW200068} and~\ref{fig:AFLW2000all}, we give the CED curves of state-of-the-art methods~\cite{feng2018joint,zhou2019CVPR2500FPS,zhu2016face,jourabloo2016large,bulat2017far} as well as RetinaFace. Even though the performance gap exists between supervised and self-supervised methods, the dense regression results of RetinaFace are comparable with these state-of-the-art methods. More specifically, we observe that (1) five facial landmarks regression can alleviate the training difficulty of dense regression branch and significantly improve the dense regression results. (2) using single-stage features (as in RetinaFace) to predict dense correspondence parameters is much harder than employing (Region of Interest) RoI features (as in Mesh Decoder~\cite{zhou2019CVPR2500FPS}). 
As illustrated in Fig.~\ref{fig:denseanalysis}, RetinaFace can easily handle faces with pose variations but has difficulty under complex scenarios. This  indicates that mis-aligned and over-compacted feature representation ($1\times1\times256$ in RetinaFace) impedes the single-stage framework achieving high accurate dense regression outputs.
Nevertheless, the projected face regions in the dense regression branch still have the effect of attention~\cite{wang2017faceattention} which can help to improve face detection as confirmed in the section of ablation study.

\subsection{Face Recognition Accuracy}
Face detection plays a crucial role in robust face recognition but its effect is rarely explicitly measured. In this paper, we demonstrate how our face detection method can boost the performance of a state-of-the-art publicly available face recognition method, \ie ArcFace~\cite{deng2018arcface}.
ArcFace~\cite{deng2018arcface}  studied how different aspects in the training process of a deep convolutional neural network (\ie, choice of the training set, the network and the loss function) affect large scale face recognition performance. However, ArcFace paper did not study the effect of face detection by applying only the MTCNN~\cite{zhang2016joint} for detection and alignment. In this paper, we replace MTCNN by RetinaFace to detect and align all of the training data (\ie MS1M~\cite{guo2016ms}) and test data (\ie LFW~\cite{huang2007labeled}, CFP-FP~\cite{sengupta2016frontal}, AgeDB-30~\cite{Moschoglou2017AgeDB} and IJBC~\cite{maze2018iarpa}), and keep the embedding network (\ie ResNet100~\cite{he2016deep}) and the loss function (\ie additive angular margin) exactly the same as ArcFace. 

In Tab.~\ref{tab:arcfacevalidation}, we show the influence of face detection and alignment on deep face recognition (\ie ArcFace) by comparing the widely used MTCNN~\cite{zhang2016joint} and the proposed RetinaFace. The results on CFP-FP, demonstrate that RetinaFace can boost ArcFace's  
verification accuracy from $98.37\%$ to $99.49\%$. This result shows that the performance of frontal-profile face verification is now approaching that of frontal-frontal face verification (\eg $99.86\%$ on LFW). 

\begin{table}[ht!]
\centering
\begin{tabular}{l|ccc}
\hline
Methods            & LFW  &  CFP-FP    &AgeDB-30\\
\hline
MTCNN+ArcFace~\cite{deng2018arcface}      & 99.83     & 98.37       &98.15\\
RetinaFace+ArcFace & {\bf 99.86}    & {\bf 99.49}  & {\bf 98.60}\\
\hline
\end{tabular}
\hspace{1in}
\caption{Verification performance ($\%$) of different methods on LFW, CFP-FP and AgeDB-30.}
\label{tab:arcfacevalidation}
\vspace{-4mm}
\end{table}

\begin{figure}[h]
\centering
\includegraphics[width=0.4\textwidth]{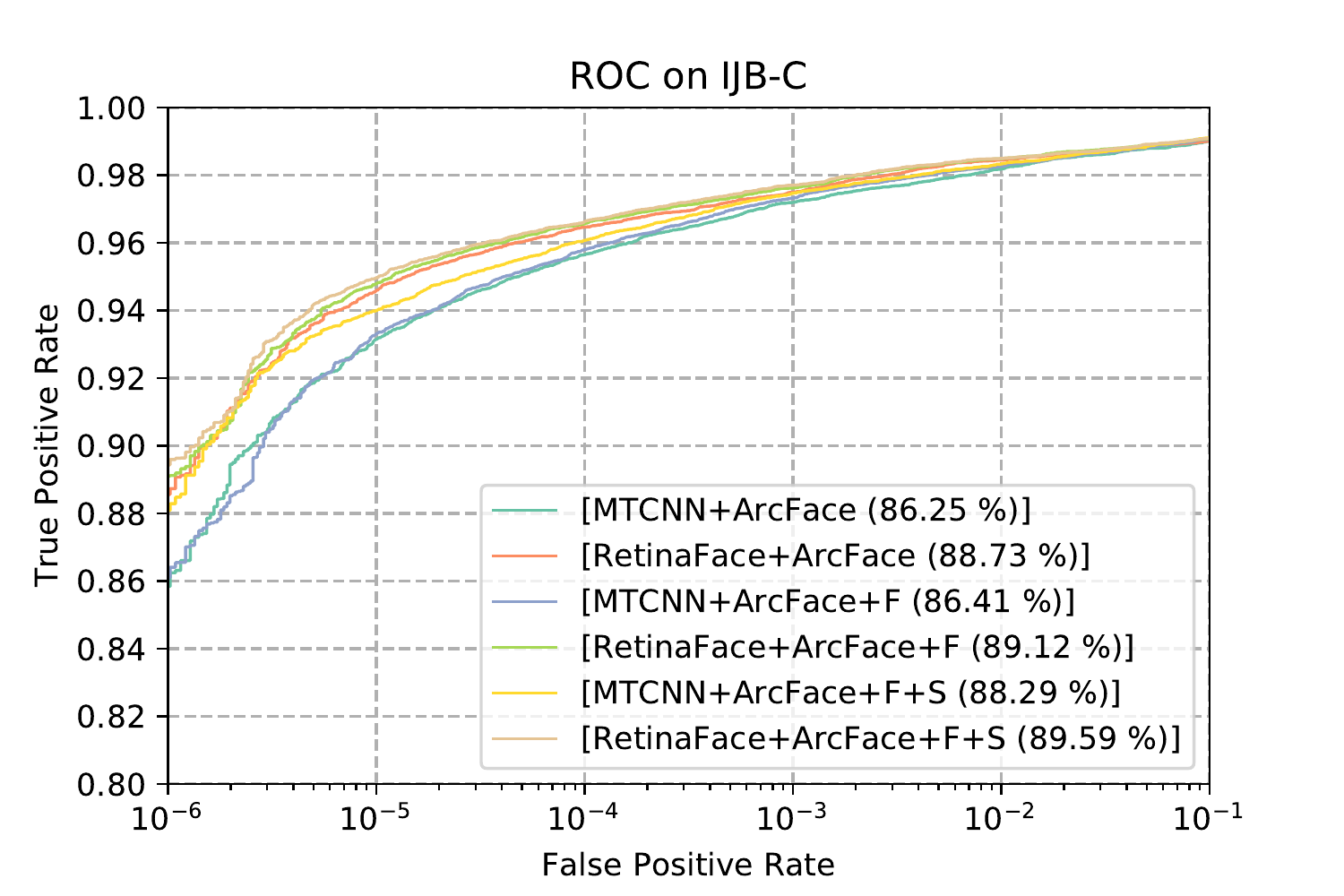}
\caption{ROC curves of 1:1 verification protocol on the IJB-C dataset. ``+F'' refers to flip test during feature embedding and ``+S'' denotes face detection score used to weigh samples within templates. We also give TAR for FAR$=1e-6$ at the end of the each legend.}
\label{fig:IJBCroc}
\vspace{-4mm}
\end{figure}

In Fig.~\ref{fig:IJBCroc}, we show the ROC curves on the IJB-C dataset as well as the TAR for FAR$=1e-6$ at the end of each legend. We employ two tricks (\ie flip test and face detection score to weigh samples within templates) to progressively improve the face verification accuracy. Under fair comparison, TAR (at FAR$=1e-6$) significantly improves from $88.29\%$ to $89.59\%$ simply by replacing MTCNN with RetinaFace. This  indicates that (1) face detection and alignment significantly affect face recognition performance and (2) RetinaFace is a much stronger baseline that MTCNN for face recognition applications.

\subsection{Inference Efficiency}

During testing, RetinaFace performs face localisation in a single stage, which is flexible and efficient.
Besides the above-explored heavy-weight model (ResNet-152, size of 262MB, and AP $91.8\%$ on the WIDER FACE hard set), 
we also resort to a light-weight model (MobileNet-0.25~\cite{howard2017mobilenets}, size of 1MB, and AP $78.2\%$ on the WIDER FACE hard set) to accelerate the inference. 

For the light-weight model, we can quickly reduce the data size by using a $7 \times 7$ convolution with stride=4 on the input image,
tile dense anchors on $P_3$, $P_4$ and $P_5$ as in~\cite{najibi2017ssh}, and remove deformable layers. In addition, the first two convolutional layers initialised by the ImageNet pre-trained model are fixed to achieve higher accuracy. 

Tab.~\ref{tab:efficiency} gives the inference time of two models with respect to different input sizes. We omit the time cost on the dense regression branch, thus the time statistics are irrelevant to the face density of the input image.
We take advantage of TVM~\cite{chen2018tvm} to accelerate the model inference and timing is performed on the NVIDIA Tesla P40 GPU, Intel i7-6700K CPU and ARM-RK3399, respectively.
RetinaFace-ResNet-152 is designed for highly accurate face localisation, running at 13 FPS for VGA images ($640\times480$).
By contrast, RetinaFace-MobileNet-0.25 is designed for highly efficient face localisation which demonstrates considerable real-time speed of 40 FPS at GPU for 4K images ($4096\times2160$), 20 FPS at multi-thread CPU for HD images ($1920\times1080$), and 60 FPS at single-thread CPU for VGA images ($640\times480$). Even more impressively, 16 FPS at ARM for VGA images ($640\times480$) allows for a fast system on mobile devices.

\begin{table}[ht!]
\centering
\begin{tabular}{l|ccc}
\hline
Backbones       & VGA  & HD  & 4K\\
\hline
ResNet-152 (GPU)    &    75.1     &  443.2        &  1742 \\
\hline
MobileNet-0.25 (GPU)    &  1.4       &  6.1        & 25.6\\
MobileNet-0.25 (CPU-m)  &  5.5    &  50.3     & -\\
MobileNet-0.25 (CPU-1)  &  17.2   &  130.4   & -\\
MobileNet-0.25 (ARM)  &  61.2   &  434.3   & -\\
\hline
\end{tabular}
\hspace{1in}
\caption{Inference time (ms) of RetinaFace with different backbones (ResNet-152 and MobileNet-0.25) on different input sizes (VGA@640x480, HD@1920x1080 and 4K@4096x2160). ``CPU-1'' and ``CPU-m'' denote single-thread and multi-thread test on the Intel i7-6700K CPU, respectively. ``GPU'' refers to the NVIDIA Tesla P40 GPU and ``ARM'' platform is RK3399(A72x2).}
\label{tab:efficiency}
\vspace{-4mm}
\end{table}

\section{Conclusions}
We studied the challenging problem of simultaneous dense localisation and alignment of faces of arbitrary scales in images and we proposed the first, to the best of our knowledge, one-stage solution (RetinaFace). Our solution outperforms state of the art methods in the current most challenging benchmarks for face detection. Furthermore, when RetinaFace is combined with state-of-the-art practices for face recognition it obviously improves the accuracy. The data and models have been provided publicly available to facilitate further research on the topic.

\section{Acknowledgements}
Jiankang Deng acknowledges financial support from the Imperial President's PhD Scholarship and GPU donations from NVIDIA.
Stefanos Zafeiriou acknowledges support from EPSRC Fellowship DEFORM (EP/S010203/1), FACER2VM (EP/N007743/1) and a Google Faculty Fellowship. 

{\small
\bibliographystyle{ieee}
\bibliography{egbib}
}

\end{document}